\documentclass[letterpaper]{article} 
\usepackage[draft]{aaai2026}  
\usepackage{times}  
\usepackage{helvet}  
\usepackage{courier}  
\usepackage[hyphens]{url}  
\usepackage{graphicx} 
\usepackage{natbib}  
\usepackage{caption} 
\usepackage{algorithm}
\usepackage{algorithmic}

\usepackage{newfloat}
\usepackage{listings}
\DeclareCaptionStyle{ruled}{labelfont=normalfont,labelsep=colon,strut=off} 
\floatstyle{ruled}
\newfloat{listing}{tb}{lst}{}
\floatname{listing}{Listing}
\title{Generation or Judgement? A Paradigm Perspective on LLM-Based \\ Emotion--Cause Pair Extraction in Conversation}

\author{
    Weijie Feng,
    Hongchuang Wang,
    Binbin Liu,
    Zhiyong Cheng
}

\affiliations{
    Hefei University of Technology\\
    wjfeng@hfut.edu.cn,
    hongchuangwang@mail.hfut.edu.cn,\\
    binbin.liu@hfut.edu.cn,
    jason.zy.cheng@gmail.com
}

\usepackage{lipsum}
\usepackage{todonotes}
\usepackage{enumitem}
\usepackage{tabularx}
\usepackage{bm,amsfonts,amsmath,dsfont}
\usepackage{microtype}
\usepackage{subcaption}
\usepackage{graphicx}
\usepackage{multirow,multicol}
\usepackage{makecell}
\usepackage{booktabs}
\usepackage{dashrule}
\usepackage{threeparttable}
\usepackage[most]{tcolorbox}
\usepackage{xcolor}
\usepackage{colortbl}
\usepackage{xspace}
\usepackage{xfp}
\usepackage{utfsym}
\usepackage{soul}

\definecolor{mygreen}{HTML}{4DAF4A}
\definecolor{myred}{HTML}{E41A1C}
\definecolor{paleblue}{HTML}{EBF5FF}
\definecolor{goodblue}{HTML}{750014}

\definecolor{1st}{HTML}{EBF5FF}

\newcommand{\gain}[2]{%
    {\setlength{\fboxsep}{0.3pt}%
    \scriptsize\colorbox{1st!#1}{\strut #2}}%
}

\newcommand{\circnum}[1]{
\tikz[baseline=-0.75ex]{
\node[
    circle,
    draw=black,
    fill=black,
    text=white,
    font=\bfseries\scriptsize,
    inner sep=0.8pt
] (n) {#1};
}}

\newtcolorbox{takeaway}{
    enhanced,
    breakable,
    colback=white,
    colframe=goodblue,
    boxrule=1.2pt,
    arc=2mm,
    left=1mm,
    right=1mm,
    top=1mm,
    bottom=1mm
}

\newcommand{\first}[1]{{\cellcolor{1st}{\textbf{#1}}}}

\definecolor{placeholdercolor}{HTML}{F44336}
\newcommand{\placeholder}[1]{%
    \textcolor{placeholdercolor}{[#1]}%
}

\newtcolorbox{promptbox}[1]{
    enhanced,
    breakable,
    colback=gray!4,
    colframe=gray!55,
    boxrule=0.5pt,
    arc=4pt,
    left=5pt,
    right=5pt,
    top=5pt,
    bottom=5pt,
    title=\textbf{#1},
    coltitle=black,
    colbacktitle=gray!12,
    fonttitle=\small,
    before upper={\small\ttfamily\raggedright},
}

\newcommand{\promptlabel}[1]{{\bfseries #1}}

\begin{document}

\maketitle

\begin{abstract}

Emotion--cause pair extraction in conversation (ECPEC) identifies utterance pairs in which one utterance causes an emotion expressed in another.
Recent LLM-based approaches formulate ECPEC at markedly different granularities, ranging from generating complete pair sets to judging individual candidate pairs.
In this paper, we make the surprising observation that task formulation substantially affects performance, where pair-level judgement outperforms dialogue-level generation in all 18 controlled comparisons.
We investigate the sources of this paradigm gap and find that many relations omitted by dialogue-level generation remain recognizable under explicit pair queries, under which the model recognizes 92.7\%--98.1\% of emotion--cause relations.
This suggests that LLMs can recognize emotion--cause relations but struggle to discover and return complete pair sets.
Pair-level judgement alleviates this burden, although its candidate rankings are more reliable than the binary decisions produced by a shared threshold.
Based on this diagnosis, we introduce an auxiliary retriever that selectively re-examines ambiguous boundary cases, yielding consistent F1 improvements of 0.50--1.46 points across three datasets while maintaining an inference time of only 1.49$\times$ that of the baseline paradigm.
These findings show that task decomposition and candidate scope are critical to effectively utilizing LLMs for ECPEC.
    
\end{abstract}
\section{Introduction}\label{sec:intro}

\begin{figure}[!t]
\centering
    \includegraphics[width=\columnwidth]{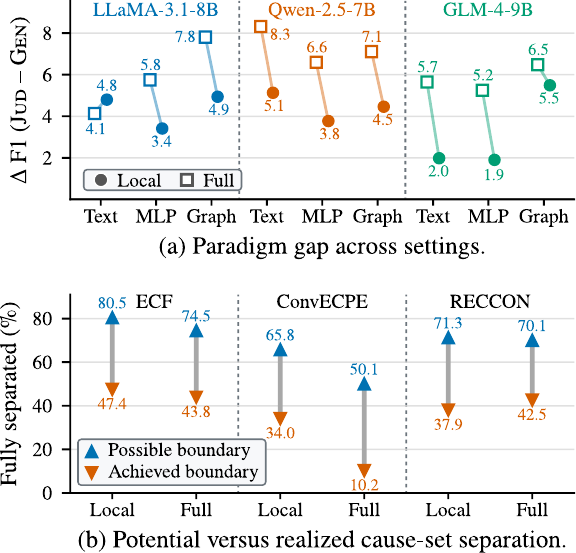}
    \caption{Paradigm gap and unrealized separation potential.
    \textsc{Jud} consistently outperforms \textsc{Gen}, while the shared decision boundary realizes only part of the separation permitted by candidate rankings.}
    \label{fig:evidence}
\end{figure}

Emotion--cause pair extraction in conversation (ECPEC) aims to identify emotion--cause utterance pairs, where a cause utterance triggers or explains the emotion conveyed by a target utterance~\cite{li2023,wang2023}.
ECPEC may involve multiple speakers and multiple causes, requiring causal reasoning over conversational context.
Prior studies~\cite{jeong2023,liu2023,an2023,wang2024,ma2026} have primarily addressed this task using task-specific architectures that explicitly model conversational context, speaker interactions, and emotion--cause dependencies.
Recently, LLMs have demonstrated strong language understanding and contextual reasoning capabilities~\cite{huang-chang-2023-towards,ijcai2024p0917,lou2024large}, which provide a natural foundation for emotion--cause reasoning.
Thus, recent studies~\cite{luo2024nusemo,ju2025,wang2026,wu2026} have increasingly explored LLM-based approaches to ECPEC.

Although LLM-based ECPEC is still in its early stages, existing studies have explored markedly different ways of using LLMs.
Tu et al.~\cite{tu2026}, for example, evaluate GPT-4 by asking it to predict the emotions and causes of all utterances from the complete conversation, whereas Luo et al.~\cite{luo2024nusemo} and Arefa et al.~\cite{arefa2024jmi} decompose the task into emotion recognition followed by target-specific cause prediction.
Ju et al.~\cite{ju2025} formulate each target--candidate pair as a yes-or-no question and use GPT-3.5 to refine the candidate decision.
More elaborate systems embed LLMs into a speaker-centric cognition--perception--reasoning agent~\cite{wang2026} or use multi-LLM debate to resolve emotion and pair predictions disputed by generative and discriminative models~\cite{wu2026}.
Despite this diversity, existing studies have not converged on a common way of posing ECPEC to LLMs, leaving open how to utilize their capabilities effectively for the task.

\begin{figure*}[!t]
\centering
    \includegraphics[width=\textwidth]{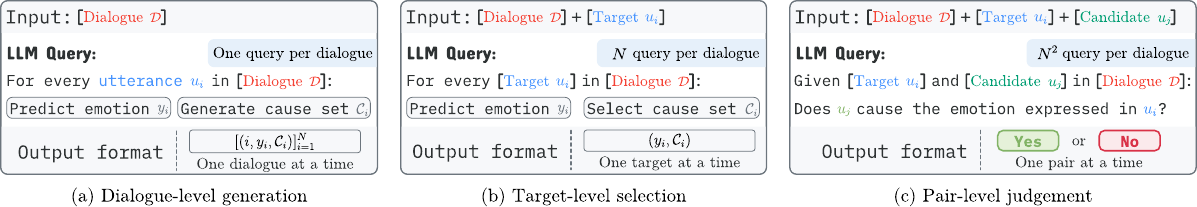}
    \caption{Illustration of the three LLM-based ECPEC paradigms.}
    \label{fig:illustration}
\end{figure*}

To explore this open question, we rethink LLM-based ECPEC from a paradigm perspective.
We distill existing approaches, according to their task-decomposition granularity, into three representative paradigms: dialogue-level generation (\textsc{Gen}), target-level selection (\textsc{Sel}), and pair-level judgement (\textsc{Jud}) (\S\ref{sec:background}).
Surprisingly, we find that paradigm choice leads to substantial differences in ECPEC performance, with \textsc{Jud} consistently outperforming \textsc{Gen} across LLM backbones and input modalities (as shown in Figure~\ref{fig:evidence}(a)),  demonstrating the broad generality of this paradigm gap (\S\ref{sec:3}).

To understand what drives this gap, we then delve into further analysis to examine how task decomposition affects the recognition of emotion--cause pairs and the recovery of complete cause sets (\S\ref{sec:4}--\S\ref{sec:judgement-boundary}).
We find that many pairs omitted by \textsc{Gen} remain recognizable under more explicit queries.
More generally, when directly asked whether a specified utterance causes the emotion in another, the LLM recognizes 92.7\%--98.1\% of gold emotion--cause pairs.
Moreover, \textsc{Gen} misses more pairs when a dialogue contains more gold pairs, a target utterance has multiple causes, or the cause is farther from the target.
These findings suggest that \textsc{Gen} struggles mainly because it must discover and return all pairs at once, rather than because the LLM cannot recognize them.
\textsc{Jud} reduces this burden by asking the LLM to examine one candidate at a time, but combining these decisions into a complete cause set remains difficult, where the model often scores gold causes above negative candidates, yet a single decision threshold cannot separate them reliably across targets (Figure~\ref{fig:evidence}(b)).
Overall, the paradigm gap arises because finer-grained task decomposition better exposes the pair-recognition ability of LLMs, but this ability is more reliably reflected in candidate rankings than in binary judgements.

Guided by these findings, we develop a lightweight remedy for the judgement paradigm that uses an auxiliary retriever to selectively re-examine ambiguous cases around the decision boundary.
The remedy consistently improves F1 by 0.50--1.46 points across three datasets, showing that adapting decision boundaries is a viable improvement direction.
Finally, our efficiency analysis shows that restricting pair-level queries to the local scope improves macro F1 over \textsc{Gen}-Full by 14.47 points while limiting inference time to 1.33$\times$, compared with 9.08$\times$ for \textsc{Jud}-Full.
Our findings show that finer-grained task decomposition, combined with a controlled candidate scope can better harness the reasoning capability of LLMs for ECPEC task while maintaining a practical balance between effectiveness and efficiency.

\section{Background}\label{sec:background}

This work focuses on the task of \textit{emotion--cause pair extraction in conversation} (ECPEC).
Given a multi-turn dialogue $\mathcal{D}=\{(u_i,s_i,\bm{x}_i)\}_{i=1}^{N}$, where $u_i$ denotes the $i$-th utterance and $s_i$ is speaker corresponding to $u_i$, $\bm{x}_i$ contains textual features and, when available, acoustic and visual features.
The goal of ECPEC is to identify all emotion--cause pairs in $\mathcal{D}$:
\begin{equation}
    \mathcal{P}^{*} = \big\{ (u_e,u_c) \mid u_c \text{ causes the emotion in } u_e \big\},
\end{equation}
Notably, an emotion utterance may have multiple causes and may also serve as its own cause.
Encouraged by the strong reasoning capabilities of LLMs, recent ECPEC studies have increasingly explored LLM-based solutions.
We distill LLM-based ECPEC approaches into three paradigms according to their inference granularity (Figure~\ref{fig:illustration}), namely dialogue-level generation, target-level selection, and pair-level judgement.

\paragraph{Dialogue-level generation.}
Dialogue-level generation (\textsc{Gen}) takes the complete dialogue $\mathcal{D}$ as input and uses a single LLM call to jointly predict an emotion label and a set of cause utterances for every utterance $u_i\in\mathcal{D}$:
\begin{equation}
    \big[ (i,y_i,\mathcal{C}_i) \big]_{i=1}^{N} = f^{\text{\textsc{Gen}}}(\mathcal{D}),
\end{equation}
where $y_i$ is the predicted emotion label of $u_i$, $\mathcal{C}_i$ denotes the set of utterance indices predicted to cause the emotion expressed in $u_i$, and $f^{\text{\textsc{Gen}}}$ is the LLM inference function.

\paragraph{Target-level selection.}
Target-level selection (\textsc{Sel}) decomposes the prediction for a dialogue $\mathcal{D}$ containing $N$ utterances into $N$ utterance-conditioned LLM calls.
Each call fixes one utterance $u_i$ as the target, treats each eligible utterance $u_j\in\mathcal{D}$, including $u_i$ itself, as a candidate cause, and predicts the emotion label $y_i$ and cause-index set $\mathcal{C}_i$:
\begin{equation}
    (y_i,\mathcal{C}_i) = f^{\text{\textsc{Sel}}}(\mathcal{D},u_i).
\end{equation}

\paragraph{Pair-level judgement.}
Pair-level judgement (\textsc{Jud}) further decomposes the target-conditioned prediction in \textsc{Sel} into independent binary decisions over individual target--candidate pairs.
For each target utterance $u_i$ and each eligible candidate utterance $u_j$, it uses a separate LLM call to determine whether $u_j$ causes the emotion expressed in $u_i$:
\begin{equation}
    f^{\text{\textsc{Jud}}}(\mathcal{D},u_i,u_j)
    \in \{\texttt{yes},\texttt{no}\}.
\end{equation}
Although \textsc{Jud} theoretically incurs $\mathcal{O}(N^2)$ LLM calls by exhaustively enumerating all target--candidate pairs in $\mathcal{D}$, our later analysis shows that restricting the candidate causes for each target $u_i$ to $\{u_j\!\mid\!\max(1,i\!-\!4)\!\leq\!j\!\leq\!i\}$ preserves performance while reducing the number of calls to $\mathcal{O}(N)$.

\section{How Does Task Paradigm Shape ECPEC?} \label{sec:3}

This section first compares the three LLM-based ECPEC paradigms against representative ECPEC methods and examines the performance differences among the paradigms (\S\ref{ssec:overall-performance}).
We then investigate whether these paradigm-level differences persist across different experimental settings (\S\ref{ssec:gap-generality}).

\subsection{Experimental Setup}
We evaluate the three paradigms on ECF~\cite{wang2023}, ConvECPE~\cite{li2023}, and RECCON~\cite{poria2021}.
LLaMA-3.1-8B~\cite{llama3} serves as the primary backbone for all three paradigms, while Qwen-2.5-7B~\cite{qwen2.5} and GLM-4-9B~\cite{glm4} are included to assess backbone generality.
Unless otherwise stated, all experiments use text-only inputs.
A key experimental setting is the candidate scope, which determines the range of utterances considered as candidate causes for each target emotion utterance.
Across the three datasets, 87.0\%--93.9\% of gold cause utterances are either the target utterance itself or one of its four preceding utterances, as detailed in Appendix Table~\ref{tab:scope-statistics}.
We therefore conduct all experiments under two candidate scopes: the local scope considers utterances $u_j$ satisfying $0\!\leq\!i-j\!\leq\!4$ for each target $u_i$, and the full scope considers every utterance in the dialogue, including both preceding and following utterances.
Further details on experiment protocols are provided in Appendix~\ref{apx:protocol}.

\begin{table}[!t]
    \centering
    \scriptsize
    \setlength{\tabcolsep}{2pt}
    \begin{tabular*}{\columnwidth}{l@{\extracolsep{\fill}}ccccccccc}
        \toprule
        & \multicolumn{3}{c}{\textbf{ECF}}
        & \multicolumn{3}{c}{\textbf{ConvECPE}}
        & \multicolumn{3}{c}{\textbf{RECCON}} \\
        \cmidrule(lr){2-4} \cmidrule(lr){5-7} \cmidrule(lr){8-10}
        \textbf{Method} & P & R & F1 & P & R & F1 & P & R & F1 \\
        \midrule
        \rowcolor{gray!25} \multicolumn{10}{c}{\textit{Conventional Method}} \\
        \midrule
        \textbf{MRC} & 44.46 & 57.65 & 50.20 & 25.92 & 47.05 & 33.42 & 52.19 & 52.86 & 52.47 \\
        \textbf{CENTER} & 47.90 & 43.65 & 44.75 & 37.98 & 45.60 & 41.44 & 47.39 & 46.88 & 47.13 \\
        \textbf{SCALE} & 55.01 & \underline{60.67} & 57.70 & 40.31 & 51.74 & 45.32 & 56.31 & \underline{61.60} & \underline{58.83} \\
        \midrule
        \rowcolor{gray!25} \multicolumn{10}{c}{\textit{Prior LLM-based Methods}} \\
        \midrule
        \textbf{DEC-Debate}$^\sharp$ & \first{69.51} & 55.52 & \underline{61.73} & 50.18 & 50.54 & 50.36 & - & - & - \\
        \textbf{GMEC-GPT}$^\sharp$ & 55.92 & 50.93 & 53.31 & - & - & - & - & - & - \\
        \textbf{MSPF+}$^\sharp$ & - & - & - & 50.31 & \underline{57.91} & 53.84 & 56.29 & 54.91 & 55.59 \\
        \midrule
        \rowcolor{gray!25} \multicolumn{10}{c}{\textit{Controlled LLM Paradigms}} \\
        \midrule
        \textbf{\textsc{Gen}-Full} & 60.04 & 53.82 & 56.76 & 28.45 & 23.55 & 25.77 & 50.36 & 55.01 & 52.58 \\
        \textbf{\textsc{Gen}-Local} & 64.98 & 51.31 & 57.34 & 50.37 & 31.75 & 38.95 & 56.55 & 56.42 & 56.49 \\
        \textbf{\textsc{Sel}-Full} & 64.69 & 57.61 & 60.94 & 52.62 & 55.09 & 53.83 & 52.33 & 57.22 & 54.66 \\
        \textbf{\textsc{Sel}-Local} & 66.52 & 55.15 & 60.30 & \underline{56.13} & 53.32 & \underline{54.69} & 57.67 & 53.20 & 55.34 \\
        \textbf{\textsc{Jud}-Full} & 51.46 & \first{74.32} & 60.81 & 33.27 & \first{79.48} & 46.90 & 52.01 & \first{65.93} & 58.15 \\
        \textbf{\textsc{Jud}-Local} & \underline{67.74} & 57.39 & \first{62.14} & \first{58.16} & \underline{56.92} & \first{57.53} & \first{63.60} & 54.78 & \first{58.86} \\
        \bottomrule
    \end{tabular*}
    \caption{Overall ECPEC performance comparison.
        \colorbox{1st}{\textbf{Shading}} and \underline{underlining} indicate the best and second-best results.
        $^\sharp$ denotes the results taken from the original papers.}
    \label{tab:aligned-paradigms}
\end{table}

\subsection{LLMs Are Competitive for ECPEC} \label{ssec:overall-performance}

Table~\ref{tab:aligned-paradigms} compares controlled implementations of the three LLM-based paradigms with conventional ECPEC methods and prior LLM-based methods.
The results first demonstrate that LLMs provide a competitive foundation for ECPEC.
Under the local scope, \textsc{Jud} achieves F1 scores of 62.14, 57.53, and 58.86 on ECF, ConvECPE, and RECCON, respectively, outperforming the strongest baselines in Table~\ref{tab:aligned-paradigms}, which obtain 61.73, 53.84, and 58.83.
Thus, appropriately designed LLM paradigms can achieve state-of-the-art ECPEC performance.
Performance, however, varies substantially across the three paradigms.
Under local scope, their F1 scores range from 57.34 to 62.14 on ECF, from 38.95 to 57.53 on ConvECPE, and from 55.34 to 58.86 on RECCON.
The corresponding ranges under full scope are 56.76--60.94, 25.77--53.83, and 52.58--58.15.
\textsc{Jud} achieves the highest F1 in four of the six settings.
The exceptions are ECF and ConvECPE under full scope, where \textsc{Sel} achieves 60.94 and 53.83 F1 respectively, compared with 60.81 and 46.90 for \textsc{Jud}.
The paradigm-level difference is particularly pronounced on ConvECPE, where the F1 spread reaches 18.58 points under local scope and 28.06 points under full scope.
Overall, adopting an LLM alone does not guarantee ECPEC performance, while the paradigm through which the task is learned and performed critically shapes its effectiveness.

\begin{table}[!t]
    \centering
    \scriptsize
    \setlength{\tabcolsep}{2pt}
    \begin{tabular*}{\columnwidth}{@{\extracolsep{\fill}}lcccccc@{}}
        \toprule
        & \multicolumn{2}{c}{\textbf{LLaMA-3.1-8B}}
        & \multicolumn{2}{c}{\textbf{Qwen-2.5-7B}}
        & \multicolumn{2}{c}{\textbf{GLM-4-9B}} \\
        \cmidrule(lr){2-3} \cmidrule(lr){4-5} \cmidrule(lr){6-7}
        \textbf{Input} & Local & Full & Local & Full & Local & Full \\
        \midrule
        Text & +4.80 & +4.05 & +5.13 & +8.32 & +1.98 & +5.65 \\
        Multimodal + MLP & +3.41 & +5.75 & +3.77 & +6.60 & +1.90 & +5.25 \\
        Multimodal + GraphSmile & +4.94 & +7.81 & +4.46 & +7.11 & +5.49 & +6.48 \\
        \bottomrule
    \end{tabular*}
    \caption{F1 improvement of \textsc{Jud} over \textsc{Gen} across settings on ECF.}
    \label{tab:gap-across-inputs}
\end{table}

\subsection{The Paradigm Gap Persists Across Settings}\label{ssec:gap-generality}

Using \textsc{Gen} and \textsc{Jud} as the two endpoints of the task-decomposition spectrum, we next examine whether the advantage of pair-level judgement generalizes across backbone architectures and input representations.
Using ECF, which also provides multimodal dialogue data, we evaluate three LLM backbones under three input configurations: text-only input, MLP-fused multimodal features, and GraphSmile-enhanced~\cite{li2025a} multimodal representations.
For each configuration, we report $\Delta F1=F1_{\textsc{Jud}}-F1_{\textsc{Gen}}$.
As shown in Table~\ref{tab:gap-across-inputs}, $\Delta F1$ is positive in all 18 backbone--input--scope combinations, ranging from 1.90 to 8.32 points.
The advantage persists with both multimodal input configurations and reaches 4.46--7.81 points with GraphSmile-enhanced representations.
Thus, providing additional multimodal evidence does not eliminate the performance difference between the two paradigms.
Moreover, the advantage is larger under full scope than under local scope in eight of the nine backbone--input configurations, indicating that pair-level decomposition generally becomes more beneficial as the candidate scope expands.

\begin{takeaway}
    \paragraph{Takeaway.}
    LLMs provide a competitive foundation for ECPEC, but their effectiveness varies substantially with task decomposition.
    Pair-level judgement (\textsc{Jud}) consistently outperforms dialogue-level generation (\textsc{Gen}) across different settings.
\end{takeaway}
\section{Why Does \textsc{Gen} Miss Emotion--Cause Pairs?}\label{sec:4}

This section diagnoses why \textsc{Gen} misses valid emotion--cause pairs.
We first determine whether these missed pairs are beyond the LLM's recognition ability or merely overlooked by \textsc{Gen} (\S\ref{ssec:recognizable-omissions}).
We then locate where the omissions arise in the generation process (\S\ref{ssec:4-2}), and examine the conditions under which \textsc{Gen} is prone to omission (\S\ref{ssec:incomplete-generation}).

\begin{figure}[!t]
    \centering
    \includegraphics[width=\columnwidth]{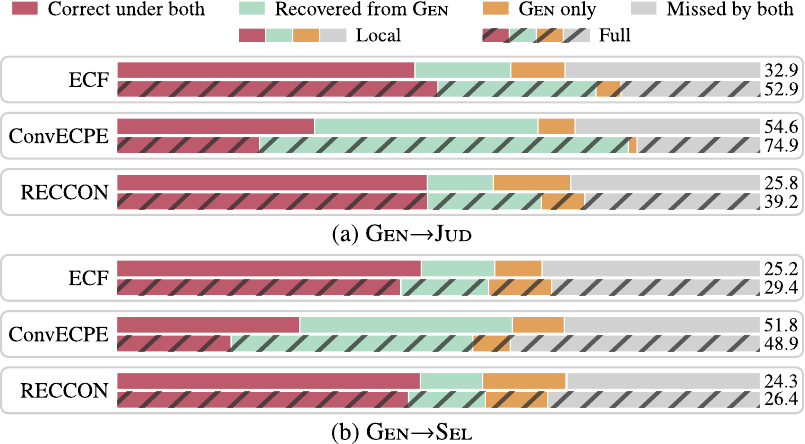}
    \caption{Emotion--cause pairs omitted by \textsc{Gen}.}
    \label{fig:same-pair-recovery}
\end{figure}

\subsection{\textsc{Gen} Omits Pairs It Can Recognize}\label{ssec:recognizable-omissions}

We ask whether the emotion--cause pairs omitted by \textsc{Gen} are beyond LLM recognition or remain recoverable under finer-grained task paradigms.
Section~\ref{sec:3} shows that \textsc{Sel} and \textsc{Jud} recognize substantially more gold pairs than \textsc{Gen}, but aggregate recall does not reveal whether their additional true positives correspond to the emotion--cause pairs omitted by \textsc{Gen}.
We therefore use both paradigms as references to re-examine the omissions of \textsc{Gen}.
To evaluate this, we define the recoverable omission rate (ROR, see Appendix~\ref{apx:recoverable-omission}) of \textsc{Gen} as the proportion of gold pairs missed by \textsc{Gen} but correctly identified by another paradigm.
Figure~\ref{fig:same-pair-recovery} partitions the gold emotion--cause pairs according to whether they are recognized by \textsc{Gen} and the corresponding reference paradigm, with the value to the right of each bar reporting the ROR.
Across the six dataset--scope conditions, \textsc{Jud} recovers 25.8--74.9\% of the gold pairs omitted by \textsc{Gen}, whereas \textsc{Sel} recovers 24.3--51.8\%.
The ROR is higher with \textsc{Jud} as the reference in all six conditions.
The highest RORs occur on ConvECPE, where \textsc{Jud} recovers 54.6\% of the gold pairs omitted by \textsc{Gen} under local scope and 74.9\% under full scope.
In the latter setting, \textsc{Jud} recovers 1,097 of the 1,464 gold pairs missed by \textsc{Gen}, whereas \textsc{Gen} itself correctly identifies only 451 gold pairs.
These results show that many emotion--cause pairs omitted by \textsc{Gen} can still be identified by LLMs in practice, but are not reliably exposed through holistic generation.


\begin{figure}[!t]
    \centering
    \includegraphics[width=\columnwidth]{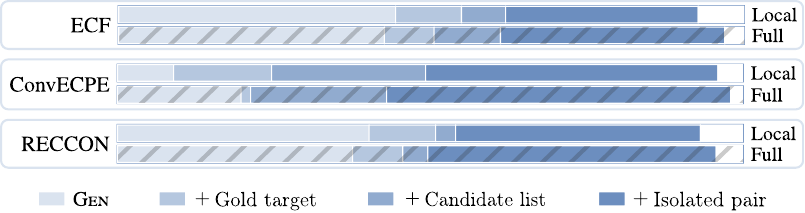}
    \caption{Results under progressive task decomposition.}
    \label{fig:progressive-disclosure}
\end{figure}

\subsection{Omissions Arise During Set-Level Prediction} \label{ssec:4-2}

We next locate where these omissions arise by progressively exposing the four key decisions hidden within \textsc{Gen}.
The first asks the model to generate all emotion--cause pairs directly from the dialogue.
The second provides a gold emotion target and its emotion label, removing target discovery and emotion-label prediction.
The third additionally enumerates all candidate causes, replacing implicit cause search with candidate-set prediction.
The final condition presents one target--candidate pair at a time, removing the need to jointly select and return a complete cause set.
We probe the same LLaMA-3.1-8B checkpoint, which is jointly fine-tuned on the three gold-target formats, with dialogue generation evaluated as an additional probe.
As shown in Figure~\ref{fig:progressive-disclosure}, providing the gold target raises pair recall by 1.51--15.67 points across the six dataset--scope conditions, while listing the candidate utterances adds another 3.20--24.67 points.
These gains suggest that target and candidate discovery account for part of the omissions.
The largest increase, however, consistently occurs when candidate-set prediction is decomposed into isolated pair judgements, improving recall by 30.83--55.04 points and yielding final recall of 92.72\%--98.07\%.
This pattern locates the dominant omission bottleneck in jointly selecting and returning a complete cause set.

\begin{table}[!t]
    \centering
    \scriptsize
    \setlength{\tabcolsep}{2pt}
    \begin{tabular*}{\columnwidth}{l@{\extracolsep{\fill}}ccccccccc}
        \toprule
        & \multicolumn{3}{c}{\textbf{ECF}}
        & \multicolumn{3}{c}{\textbf{ConvECPE}}
        & \multicolumn{3}{c}{\textbf{RECCON}} \\
        \cmidrule(lr){2-4} \cmidrule(lr){5-7} \cmidrule(lr){8-10}
        \textbf{Metric} & Low & Mid & High & Low & Mid & High & Low & Mid & High \\
        \midrule
        \#Dialogues & 88 & 77 & 79 & 10 & 11 & 10 & 62 & 92 & 71 \\
        Avg. \#Gold Pairs & 2.5 & 6.6 & 14.5 & 37.4 & 58.9 & 89.3 & 4.2 & 6.9 & 12.3 \\
        \midrule
        \textsc{Gen} Recall & 55.3 & 61.1 & 50.3 & 43.0 & 18.4 & 19.1 & 62.8 & 59.4 & 49.5 \\
        \textsc{Sel} Recall & 59.0 & 62.5 & 55.2 & 49.2 & 53.5 & 58.7 & 65.5 & 61.6 & 51.5 \\
        \textsc{Jud} Recall & 69.1 & 77.6 & 73.8 & 71.9 & 79.5 & 82.6 & 73.2 & 70.7 & 60.3 \\
        \textsc{Gen} P/G & 1.02 & 0.96 & 0.84 & 0.95 & 0.78 & 0.81 & 1.80 & 1.11 & 0.87 \\
        \bottomrule
    \end{tabular*}
    \caption{Performance versus the number of gold pairs.}
    \label{tab:relation-load}
\end{table}

\subsection{When Is \textsc{Gen} More Likely to Miss Pairs?}\label{ssec:incomplete-generation}

We next examine conditions under which \textsc{Gen} is more likely to miss gold emotion--cause pairs.

\paragraph{Number of gold pairs.}
We first examine whether \textsc{Gen} omits more gold pairs as the number of pairs in a dialogue increases.
Within each dataset, we partition dialogues into approximately equal-sized low-, medium-, and high-count groups according to their numbers of gold pairs.
We report recall to measure the proportion of gold pairs recovered and the predicted-to-gold ratio (P/G) to assess whether \textsc{Gen}'s output-set size grows proportionally with the gold set.
As shown in Table~\ref{tab:relation-load}, recall under full-scope in the high-count group is lower than in the low-count group by 5.0--23.9 points.
P/G likewise decreases from 1.02 to 0.84, from 0.95 to 0.81, and from 1.80 to 0.87.
These decreases indicate that, in high-count dialogues, \textsc{Gen} recovers a smaller proportion of the required pairs and produces fewer predictions relative to the gold-set size.
The cross-paradigm comparison is most revealing on ConvECPE, where \textsc{Gen} recall decreases by 23.9 points, whereas the recalls of \textsc{Sel} and \textsc{Jud} increase by 9.5 and 10.7 points, respectively.
By contrast, all three paradigms decline on RECCON, suggesting that high-count dialogues in this dataset are also intrinsically more difficult.

\begin{table}[!t]
    \centering
    \scriptsize
    \setlength{\tabcolsep}{2.5pt}
    \begin{tabular*}{\columnwidth}{l@{\extracolsep{\fill}}ccccccccc}
        \toprule
        & \multicolumn{3}{c}{\textbf{Any Cause}}
        & \multicolumn{3}{c}{\textbf{Cause Recall}}
        & \multicolumn{3}{c}{\textbf{Complete}} \\
        \cmidrule(lr){2-4} \cmidrule(lr){5-7} \cmidrule(lr){8-10}
        \textbf{\#Causes} & 1 & 2 & $\ge$ 3 & 1 & 2 & $\ge$ 3 & 1 & 2 & $\ge$ 3 \\
        \midrule
        \textsc{Gen} & 52.2 & 61.3 & 61.3 & 52.2 & 43.4 & 29.8 & 52.2 & 25.5 & 7.8 \\
        \textsc{Sel} & 67.1 & 77.4 & 78.2 & 67.1 & 55.9 & 39.3 & 67.1 & 34.3 & 7.7 \\
        \textsc{Jud} & 82.2 & 89.7 & 89.1 & 82.2 & 74.3 & 54.9 & 82.2 & 58.8 & 23.2 \\
        \bottomrule
    \end{tabular*}
    \caption{Cause recovery (\%) versus the number of causes.}
    \label{tab:cause-multiplicity}
\end{table}

\paragraph{Number of gold causes.}
We next examine whether \textsc{Gen} becomes less complete as the number of gold causes per target increases.
We group targets by cause count and report whether at least one cause is recovered (Any Cause), the proportion recovered (Cause Recall), and whether all causes are recovered (Complete).
Table~\ref{tab:cause-multiplicity} reports the average result for three datasets under full-scope, \textsc{Gen}'s any-cause recovery remains stable, while its cause recall falls from 52.2\% to 29.8\% and complete recovery from 52.2\% to 7.8\%.
Among targets with at least three causes, it recovers some but not all causes in 53.5\% of cases, indicating difficulty in constructing a complete cause set.
This problem affects all three paradigms, although \textsc{Jud} mitigates it with 54.9\% cause recall and 23.2\% complete recovery in the same group.


\begin{figure}[!t]
    \centering
    \includegraphics[width=\columnwidth]{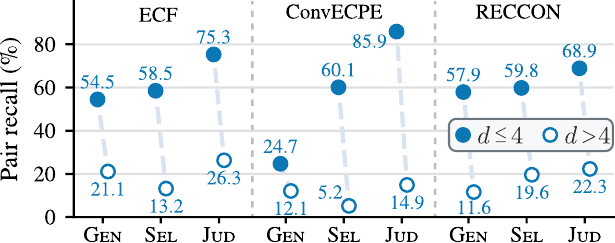}
    \caption{Pair recall (\%) versus emotion--cause distances.}
    \label{fig:distance-controlled-recall}
\end{figure}

\paragraph{Emotion--cause distance.}
Finally, we examine whether \textsc{Gen} is more likely to miss a gold pair when its emotion target and cause are farther apart.
We report pair recall for local ($d\leq4$) and long-range ($d>4$) gold pairs under full-scope, where $d$ denotes the utterance-index distance.
As shown in Figure~\ref{fig:distance-controlled-recall}, \textsc{Gen}'s recall decreases from local to long-range pairs by 33.4, 12.6, and 46.3 points on the three datasets.
Long-range pairs are therefore more difficult for \textsc{Gen}, as they are for the other two paradigms.
However, even among local pairs, \textsc{Jud} outperforms \textsc{Gen} by 20.8, 61.2, and 11.0 points across the three datasets.
Thus, long distance aggravates \textsc{Gen}'s omissions but cannot explain them alone, as substantial omissions persist even when the cause is nearby.

\begin{takeaway}
    \textbf{Takeaway.}
    LLMs can recognize emotion--cause pairs, but \textsc{Gen} omits many during set-level prediction, especially with more gold pairs per dialogue, more gold causes per target, or longer emotion--cause distances.
\end{takeaway}
    
\section{Why Does \textsc{Jud} Work? When Does It Fail?}\label{sec:judgement-boundary}

This section first briefly summarizes why pair-level \textsc{Jud}, at the opposite end of the task-decomposition spectrum from dialogue-level \textsc{Gen}, is more effective (\S\ref{ssec:why-judgement-works}).
We then investigate where \textsc{Jud} reaches its limits from two perspectives.
We first examine whether \textsc{Jud} correctly ranks and separates gold causes from negative candidates within a given candidate scope, and then analyze how its pair-level errors accumulate as the scope expands (\S\ref{ssec:when-judgement-fails}).

\subsection{Explicit Pair Query Reduce Extraction Burden}\label{ssec:why-judgement-works}

\textsc{Gen} must discover emotion targets and their causes before jointly selecting and returning a complete pair set.
By explicitly specifying both the target and candidate cause, \textsc{Jud} instead reduces each query to judging one explicit emotion--cause pair.
As shown in Table~\ref{tab:aligned-paradigms} and Figure~\ref{fig:progressive-disclosure}, many gold pairs omitted under coarser paradigms are recovered when presented individually to the same checkpoint.
Together with the analysis in Section~\ref{sec:4}, these results show that \textsc{Jud} is effective because it removes the set-level prediction bottleneck and more directly elicits the LLM's pair-recognition capability.

\subsection{The Capability Boundary of \textsc{Jud}}\label{ssec:when-judgement-fails}

Expanding the candidate scope from local to full covers more gold pairs and improves \textsc{Jud}'s recall, but substantially lowers its precision and ultimately reduces F1.
This recall--precision trade-off suggests that the advantage of \textsc{Jud} reaches a limit.
We locate this boundary from two perspectives: whether gold causes are correctly ranked above and separated from negatives, and how pair-level errors accumulate as the candidate space expands.

\begin{table}[!t]
    \centering
    \scriptsize
    \setlength{\tabcolsep}{2pt}
    \begin{tabular*}{\columnwidth}{l@{\extracolsep{\fill}}cccccc}
        \toprule
        & \multicolumn{2}{c}{\textbf{ECF}}
        & \multicolumn{2}{c}{\textbf{ConvECPE}}
        & \multicolumn{2}{c}{\textbf{RECCON}} \\
        \cmidrule(lr){2-3} \cmidrule(lr){4-5} \cmidrule(lr){6-7}
        \textbf{Outcome} & Local & Full & Local & Full & Local & Full \\
        \midrule
        Case-1 & 60.33 & 74.31 & 59.70 & 79.55 & 55.90 & 68.98 \\
        Case-2 & 2.84 & 2.12 & 6.29 & 1.87 & 4.10 & 2.18 \\
        Case-3 & 29.32 & 20.29 & 20.33 & 16.21 & 27.86 & 23.61 \\
        Case-4 & 7.51 & 3.27 & 13.69 & 2.37 & 12.14 & 5.24 \\
        \midrule
        ISR & 80.48 & 74.52 & 65.80 & 50.12 & 71.31 & 70.08 \\
        RSR & 47.41 & 43.82 & 34.03 & 10.17 & 37.94 & 42.50 \\
        \bottomrule
    \end{tabular*}
    \caption{Pairwise ranking and threshold outcomes (\%).}
    \label{tab:ranking-threshold}
\end{table}

\paragraph{Ranking versus thresholding.}
To determine whether this degradation arises mainly from ranking or thresholding, we examine the judgement score and \texttt{Yes-or-No} decision assigned by \textsc{Jud} to each candidate utterance.
Reliable cause-set prediction requires every gold cause to outrank every negative candidate and the decision threshold to separate them.
For each target utterance, we therefore compare every gold cause with every negative candidate and divide the comparisons into four cases:
gold cause is accepted and the negative candidate rejected (Case-1);
gold cause scores higher, but both are accepted (Case-2);
gold cause scores higher, but both are rejected (Case-3);
or gold cause is ranked below or tied with the negative candidate (Case-4).
As shown in Table~\ref{tab:ranking-threshold}, Case-4 accounts for only 2.37\%--13.69\% of comparisons, whereas Case-2 and Case-3 together account for 18.08\%--32.16\% and are more frequent in every setting.
Case-3 also consistently dominates Case-2, showing that thresholding errors more often reject correctly ranked gold causes than accept negatives.
Thus, failures arise more often from failing to separate correctly ordered gold--negative comparisons than from ranking them incorrectly.
Pairwise comparisons, however, do not show whether all candidates for a target can be separated simultaneously.
We therefore define the Ideal Separability Ratio (ISR) as the proportion of targets whose lowest-scoring gold cause outranks their highest-scoring negative.
An oracle target-specific threshold could fully separate these targets without changing their rankings.
ISR ranges from 50.12\% to 80.48\%, whereas the Realized Separability Ratio (RSR), which measures complete separation under the threshold used by \textsc{Jud}, reaches only 10.17\%--47.41\%.
The resulting 27.58--39.95 point gap identifies target utterances whose rankings are already separable but whose cause sets are not realized by the threshold.

\begin{table}[t]
    \centering
    \scriptsize
    \setlength{\tabcolsep}{2pt}
    \begin{tabular*}{\columnwidth}{@{\extracolsep{\fill}}lccccccccc@{}}
        \toprule
        & \multicolumn{3}{c}{\textbf{ECF}}
        & \multicolumn{3}{c}{\textbf{ConvECPE}}
        & \multicolumn{3}{c}{\textbf{RECCON}} \\
        \cmidrule(lr){2-4} \cmidrule(lr){5-7} \cmidrule(lr){8-10}
        \textbf{Metric} & Local & Added & Full & Local & Added & Full & Local & Added & Full \\
        \midrule
        \#Gold & 1,758 & 115 & 1,873 & 1,666 & 249 & 1,915 & 1,655 & 112 & 1,767 \\
        \#Neg. & 8,543 & 23,026 & 31,569 & 6,139 & 86,293 & 92,432 & 8,120 & 19,346 & 27,466 \\
        TP & 1,361 & 31 & 1,392 & 1,446 & 76 & 1,522 & 1,140 & 25 & 1,165 \\
        TPR (\%) & 77.42 & 26.96 & 74.32 & 86.79 & 30.52 & 79.48 & 68.88 & 22.32 & 65.93 \\
        FP & 1,223 & 90 & 1,313 & 2,620 & 433 & 3,053 & 1,011 & 64 & 1,075 \\
        FPR (\%) & 14.32 & 0.39 & 4.16 & 42.68 & 0.50 & 3.30 & 12.45 & 0.33 & 3.91 \\
        Precision (\%) & 52.67 & 25.62 & 51.46 & 35.56 & 14.93 & 33.27 & 53.00 & 28.09 & 52.01 \\
        Recall (\%) & 72.66 & 1.66 & 74.32 & 75.51 & 3.97 & 79.48 & 64.52 & 1.41 & 65.93 \\
        \bottomrule
    \end{tabular*}
    \caption{Error accumulation under candidate-space expansion.}
    \label{tab:candidate-expansion}
\end{table}

\paragraph{Error accumulation.}
We next ask whether pair-level errors accumulate when many independently judged candidates are combined into cause sets.
We partition the candidates from the same full-scope run into Local ($0\!\leq\!i\!-\!j\!\leq\!4$) and Added ($i\!-\!j\!\geq\!5$ or $i\!-\!j\!<\!0$) regions.
As shown in Table~\ref{tab:candidate-expansion}, the Added region of \textsc{Jud} has an FPR of only 0.33\%--0.50\%, substantially below the Local FPR of 12.45\%--42.68\%.
However, the Added region contains 19,346--86,293 negatives, 2.38--14.06 times as many as the Local region, but only 112--249 gold pairs.
Consequently, this low per-pair FPR still produces 64--433 additional false positives for only 25--76 true positives.
These predictions contribute just 1.41--3.97 points to full-scope recall while lowering precision by 0.99--2.29 points.
Although 85.8\%--94.0\% of full-scope false positives originate from the harder Local region, candidate expansion introduces a secondary accumulation effect: even rare errors become consequential when independent judgements are made over a large negative space.

\begin{takeaway}
\paragraph{Takeaway.}
\textsc{Jud} is effective because it reduces ECPEC to recognizing one explicit pair at a time.
However, it remains constrained by a shared decision boundary and error accumulation as the candidate space expands.
\end{takeaway}

\section{A Diagnosis-Guided Remedy}\label{sec:remedy}

Based on the above analyses, we develop a lightweight remedy around three design choices.
\circnum{1} We adopt pair-level \textsc{Jud} rather than dialogue-level \textsc{Gen}, as explicit pair queries better expose the relation recognition capability of LLMs.
\circnum{2} We use local as the default candidate scope because it retains most gold pairs while limiting the error accumulation and inference cost caused by full-scope expansion.
\circnum{3} We introduce a localized approximation to target-specific boundary adaptation.
Directly estimating a reliable target-specific boundary is difficult, while Table~\ref{tab:ranking-threshold} shows that 16.21\%--29.32\% of instances fall into Case-3, where gold causes are ranked above negative candidates yet both are rejected.
This prevalent under-selection motivates a localized correction that reconsiders only the highest-ranked rejected candidate.
We implement this correction using an auxiliary retriever initialized from the trained \textsc{Jud} checkpoint and further fine-tuned on a mixture of original pair-judgement instances and boundary-candidate instances.
At inference, the retriever re-examines the highest-ranked rejected candidate.
If accepted, the candidate is added to the original prediction; otherwise, the prediction remains unchanged.

\begin{table}[!t]
    \centering
    \scriptsize
    \setlength{\tabcolsep}{2.0pt}
    \begin{tabular*}{\columnwidth}{l@{\extracolsep{\fill}}lll}
        \toprule
        \textbf{Metric} & \textbf{ECF} & \textbf{ConvECPE} & \textbf{RECCON} \\
        \midrule
        \textsc{Jud} F1         & 62.14 & 57.53 & 58.86 \\
        \quad + Retriever F1    & \textbf{63.60} \gain{146}{(+1.46)} & \textbf{58.03} \gain{50}{(+0.50)}    & \textbf{59.39} \gain{53}{(+0.53)} \\
        $\Delta$Precision       & -2.50 & -0.93 & -3.13 \\
        $\Delta$Recall          & +4.65 & +1.93 & +3.57 \\
        Correct additions rate  & 44.9  & 38.9  & 34.4  \\
        \bottomrule
    \end{tabular*}
    \caption{Effectiveness of the proposed remedy under local scope.
    Correct additions rate denotes the percentage of newly accepted pairs that are gold pairs.}
    \label{tab:remedy-effectiveness}
\end{table}

\paragraph{Effectiveness.}

We evaluate the final correction against its pipeline-aligned \textsc{Jud}-Local baseline for each dataset.
As shown in Table~\ref{tab:remedy-effectiveness}, the correction yields modest but consistent Pair-F1 improvements of 0.50--1.46 points across the three datasets, increasing macro F1 by 0.83 points.
The gains are recall-driven: recall increases by 1.93--4.65 points, while precision decreases by 0.93--3.13 points.
Among the candidates newly accepted by the retriever, 34.4\%--44.9\% are gold causes.
Although the accepted additions contain false positives, the recovered gold causes are sufficient to produce a net F1 improvement on every dataset.
The overall gains remain incremental because the correction reconsiders only one rejected boundary candidate per target and recovers only 11.71\%--19.12\% of the gold causes presented for reconsideration.
Nevertheless, the consistent improvements obtained without changing the original candidate rankings show that the shared decision boundary is an actionable source of error.
More comprehensive estimation of target-specific decision boundaries therefore represents a promising direction for future ECPEC research.

\begin{table}[!t]
    \centering
    \scriptsize
    \setlength{\tabcolsep}{2pt}
    \begin{tabular*}{\columnwidth}{l@{\extracolsep{\fill}}cccc}
        \toprule
        \textbf{Configuration} & \textbf{Queries} & \textbf{Input tokens}
        & \textbf{Output tokens} & \textbf{Time} \\
        \midrule
        \textsc{Gen}-Full & 1.0 & 1.00$\times$ & 1.00$\times$ & 1.00$\times$ \\
        \textsc{Gen}-Local& 1.0 & 1.03$\times$ & 0.99$\times$ & 0.98$\times$ \\
        \textsc{Sel}-Full & 12.8 & 22.00$\times$ & 1.27$\times$ & 1.39$\times$ \\
        \textsc{Sel}-Local& 12.8 & 22.32$\times$ & 1.25$\times$ & 1.36$\times$ \\
        \textsc{Jud}-Full & 303.7 & 792.27$\times$ & 2.16$\times$ & 9.08$\times$ \\
        \textsc{Jud}-Local & 53.9 & 88.90$\times$ & 0.38$\times$ & 1.33$\times$ \\
        \quad + Retriever & 66.2 & 111.19$\times$ & 0.38$\times$ & 1.49$\times$ \\
        \bottomrule
    \end{tabular*}
    \caption{Inference-cost comparison of the selected configurations.
    Input tokens, output tokens, and inference time are normalized to \textsc{Gen}-Full.}
    \label{tab:paradigm-efficiency}
\end{table}

\paragraph{Inference efficiency.}

Beyond effectiveness, another practical concern is the inference cost of the three paradigms and the proposed remedy.
Table~\ref{tab:paradigm-efficiency} compares their inference costs.
Although \textsc{Jud}-Full incurs substantially higher costs than \textsc{Gen}-Full, restricting candidate causes to the local scope reduces its inference time from 9.08$\times$ to 1.33$\times$.
Notably, despite processing 88.90$\times$ more input tokens, \textsc{Jud}-Local consumes only 0.38$\times$ as many output tokens because each query requires only a short binary response.
The retriever uses next-token classification without text generation, leaving the output-token cost unchanged.
It increases the number of queries and input tokens by 22.8\% and 25.1\% over \textsc{Jud}-Local, respectively, while adding only 11.8\% inference time.
Its total inference time remains 1.49$\times$ that of \textsc{Gen}-Full, far below the 9.08$\times$ cost of \textsc{Jud}-Full.
Thus, the remedy trades a measurable but controlled increase in inference cost for consistent F1 gains.

These gains nevertheless remain modest because the remedy performs only a localized correction rather than directly estimating a reliable target-specific decision boundary.
Its main value is therefore to demonstrate that the shared decision boundary is an actionable source of error, while more comprehensive boundary adaptation remains an open challenge.
The effectiveness and efficiency results jointly show that utilizing the existing pair-recognition ability of LLMs requires appropriate task decomposition and controlled candidate scopes, as well as decision mechanisms that can reliably translate this ability into final predictions.

\section{Related Work} \label{sec:related_work}

\paragraph{Conventional ECPEC methods.}
Emotion--cause analysis originated from emotion cause extraction, which identifies the cause of a given emotion~\cite{lee2010,gui2016}.
Xia and Ding~\cite{xia2019} later introduced emotion--cause pair extraction (ECPE) to jointly identify emotions and their causes.
Subsequent studies developed pipeline, end-to-end, graph-based, sequence-labeling, and machine-reading-comprehension methods for document-level ECPE~\cite{ding2020,chen2020,fan2021,chen2022,zhu2024}.
Poria et al.~\cite{poria2021} introduced emotion-cause recognition in conversations through RECCON, and Li et al.~\cite{li2023} subsequently formalized ECPEC as the joint extraction of emotion--cause utterance pairs and released the ConvECPE dataset.
Later ECPEC methods capture pair relations, speaker interactions, dialogue context, and event structure using mixture-of-experts, machine-reading-comprehension, and graph-based architectures~\cite{jeong2023,liu2023,an2023,wang2024}.
Multimodal extensions further incorporate acoustic and visual information and strengthen cross-modal interaction, task coupling, and global pair modeling~\cite{wang2023,li2024,hu-etal-2024-unimeec,LI2025112333,tu2026,xie2026a,ma2026}.
This line of work primarily advances task-specific architectures, whereas we study how different inference paradigms shape the use of LLMs for ECPEC.

\paragraph{LLM-based ECPEC methods.}
LLMs have recently been incorporated into ECPEC as primary predictors, auxiliary reasoners, or components of hybrid systems.
Several SemEval-2024 systems use GPT-based emotion recognition followed by a separate cause extractor~\cite{levchenko2024pweitinlp,kazakov2024petkaz}, while others use LLMs for multi-stage cause prediction or instruction-tuned joint prediction~\cite{arefa2024jmi,cheng2024mips,luo2024nusemo}.
Ju et al.~\cite{ju2025} develop a single-stage generative framework augmented with pair-level yes-or-no decisions obtained from GPT-3.5.
At the dialogue level, Tu et al.~\cite{tu2026} evaluate GPT-4 by asking it to predict emotions and causes for the complete conversation.
Wu et al.~\cite{wu2026} instead generate causes for individual targets and apply pair-level debate to disputed candidates.
More elaborate systems also embed LLMs into multi-stage agent frameworks~\cite{wang2026}.
These approaches often combine multiple inference granularities and differ substantially in backbone, modality, supervision, and candidate scope.
We therefore distill their inference operations into dialogue-level generation, target-level selection, and pair-level judgement, and compare these paradigms under controlled settings.
\section{Conclusion} \label{sec:conclusion}

This work rethinks LLM-based ECPEC from a paradigm perspective by systematically comparing dialogue-level generation, target-level selection, and pair-level judgement.
We identify a consistent paradigm gap: finer-grained task decomposition better exposes the ability of LLMs to recognize emotion--cause pairs, allowing pair-level judgement to consistently outperform dialogue-level generation.
Our analyses show that dialogue-level generation is primarily limited by the burden of discovering and returning complete pair sets, whereas pair-level judgement is limited by the difficulty of converting reliable candidate rankings into binary decisions under a shared decision boundary.
Guided by this diagnosis, we develop an auxiliary retriever that selectively re-examines ambiguous boundary cases, yielding consistent F1 improvements across all three datasets.
Overall, our analysis suggests that, among the evaluated configurations, \textsc{Jud}-Local offers the most practical foundation for LLM-based ECPEC, balancing the performance benefits of pair-level decomposition against its inference cost.
Developing reliable target-specific decision boundaries remains an important direction for translating strong candidate rankings into more accurate emotion--cause pair predictions.
We leave this direction to future work.


\bibliography{bib/ecpe,bib/ecpec,bib/erc,bib/other}


\clearpage
\newpage

\appendix
\section{Prompt Templates}\label{apx:prompts}

This section presents the prompt templates used for the three controlled paradigms.

\begin{promptbox}{Prompt for dialogue-level generation (\textsc{Gen})}
You are an expert in emotion--cause pair extraction in conversations.

Given the following conversation, identify the emotion expressed in each utterance and the utterance IDs that cause that emotion.

Emotion labels must be selected from:

\placeholder{EMOTION LABEL SET}

\promptlabel{Rules:}

1. If an utterance is neutral, set "cause\_ids" to []. 

2. If an utterance expresses a non-neutral emotion, "cause\_ids" should contain the IDs of all utterances that trigger or explain that emotion. 

3. A cause may be the emotion utterance itself. 

4. \placeholder{SCOPE INSTRUCTION}

5. Return only a valid JSON list without any explanation. 

\promptlabel{Conversation: }

\placeholder{DIALOGUE}

\promptlabel{Output format:}

[
  \{"utterance\_id": [ID],
   "emotion": "[EMOTION]",
   "cause\_ids": [[CAUSE IDS]]\}
]
\end{promptbox}

\begin{promptbox}{Prompt for target-level selection (\textsc{Sel})}
You are an expert in emotion--cause pair extraction in conversations.

Given the following conversation and target utterance, identify the emotion expressed in the target and select all utterances that cause that emotion.

Emotion labels must be selected from:

\placeholder{EMOTION LABEL SET}

\promptlabel{Rules:}

1. If the target utterance is neutral, [IDS] must be [].

2. If the target utterance expresses a non-neutral emotion, [IDS] should contain the IDs of all utterances that trigger or explain that emotion.

3. A cause may be the target utterance itself.

4. \placeholder{SCOPE INSTRUCTION}

5. Return only the filled template without any explanation.

\promptlabel{Conversation:}

\placeholder{DIALOGUE}

\promptlabel{Target utterance:}

\placeholder{TARGET ID}. \placeholder{TARGET SPEAKER}: \placeholder{TARGET TEXT}

\promptlabel{Output template:}

[TARGET TEXT] contains [CLASS] emotion, corresponding to [CAUSE IDS].
\end{promptbox}

\begin{promptbox}{Prompt for pair-level judgement (\textsc{Jud})}
You are an expert in emotion--cause pair extraction in conversations.

Given the following conversation, determine whether the candidate cause utterance triggers or explains a non-neutral emotion expressed in the target utterance.

A valid cause may be an event, statement, behavior, or contextual clue that triggers or explains the target emotion.

\promptlabel{Conversation:}

\placeholder{DIALOGUE}

\promptlabel{Target utterance:}

\placeholder{TARGET ID}. \placeholder{TARGET SPEAKER}: \placeholder{TARGET TEXT}

\promptlabel{Candidate cause utterance:}

\placeholder{CANDIDATE ID}. \placeholder{CANDIDATE SPEAKER}: \placeholder{CANDIDATE TEXT}

\promptlabel{Question:}

Does the candidate cause utterance cause a non-neutral emotion expressed in the target utterance?

Answer only "yes" or "no".

\promptlabel{Answer:}
\end{promptbox}

\noindent For local scope, \texttt{\placeholder{SCOPE INSTRUCTION}} is replaced with:
\begin{quote}
\texttt{Only consider the target utterance itself and up to four preceding utterances as possible causes.}
\end{quote}

\noindent For full scope, it is replaced with:
\begin{quote}
\texttt{Consider every utterance in the conversation as a possible cause, regardless of its position relative to the target utterance.}
\end{quote}
\section{Detailed Experimental Protocol}\label{apx:protocol}

\subsection{Datasets}

We conduct experiments on ECF~\cite{wang2023}, ConvECPE~\cite{li2023}, and RECCON~\cite{poria2021}.
These benchmarks differ substantially in dialogue length, pair density, and the number of annotated emotion--cause pairs, providing complementary evaluation conditions.
Table~\ref{tab:dataset-statistics} reports the statistics of three benchmarks.

\begin{table}[h]
\centering
\scriptsize
\setlength{\tabcolsep}{4.5pt}
\begin{tabular*}{\columnwidth}{@{\extracolsep{\fill}}lccc}
    \toprule
    \textbf{Statistic} & \textbf{RECCON} & \textbf{ConvECPE} & \textbf{ECF} \\
    \midrule
    \#Dialogue & 1,106 & 151   & 1,374 \\
    \#Utterance & 11,104 & 7,433 & 13,619 \\
    \#Emotion--Cause Pairs & 9,124 & 9,473 & 9,794 \\
    Train/Val/Test & 75/5/20 & 80/--/20 & 70/10/20 \\
    \bottomrule
\end{tabular*}
\caption{Statistics of three datasets.}
\label{tab:dataset-statistics}
\end{table}

\paragraph{ECF.}
ECF is a multimodal ECPEC benchmark derived from the television series \textit{Friends}.
It provides textual, acoustic, and visual information together with utterance-level emotion labels and emotion--cause pair annotations.
Our primary experiments use only the dialogue transcripts so that the three paradigms receive the same textual evidence.
The additional multimodal experiment uses the acoustic and visual inputs provided by the benchmark.

\paragraph{ConvECPE.}
ConvECPE is constructed from IEMOCAP~\cite{busso2008} and contains relatively long dyadic conversations.
Compared with ECF and RECCON, its test dialogues are substantially longer and therefore induce a much larger candidate space under the full scope.
This property makes ConvECPE particularly useful for examining omission under set generation and false-positive accumulation under pair-level judgement.

\paragraph{RECCON.}
RECCON is a widely used benchmark for emotion-cause pair extraction and consists of two subsets.
RECCON-DD, annotated from DailyDialog~\citep{li2017}, serves as the primary corpus for model training and evaluation, while RECCON-IE, annotated from IEMOCAP, is a smaller subset used exclusively to assess model generalization.
In this work, we choice the RECCON-DD subset only and construct each gold pair from an annotated emotion utterance and one of its annotated cause utterances.

\subsection{Candidate Scope}\label{apx:candidate-scope}

For each target emotion utterance $u_i$, the candidate scope determines which utterances are considered as potential causes.
Under local scope, the candidate set is restricted to $\{u_j\! \mid\! 0\!\leq\! i\!-\!j\!\leq\!4\}$, which includes the target itself and its four preceding utterances.
Under full scope, every utterance in the dialogue is treated as a candidate cause, including utterances occurring after the target.

\begin{table}[h]
    \centering
    \scriptsize
    \setlength{\tabcolsep}{2pt}
    \begin{tabular*}{\columnwidth}{@{\extracolsep{\fill}}lrrr}
        \toprule
        \textbf{Statistic} & \textbf{ECF} & \textbf{ConvECPE} & \textbf{RECCON-DD} \\
        \midrule
        \#Gold Pairs & 1,873 & 1,915 & 1,767 \\
        Mean $|d|$ & 0.83 & 1.85 & 1.38 \\
        $0\leq d\leq4$ & 93.9\% & 87.0\% & 93.7\% \\
        \bottomrule
    \end{tabular*}
    \caption{Utterance-distance statistics of gold emotion--cause pairs, where $d=i-j$ for target utterance $u_i$ and cause utterance $u_j$.}
    \label{tab:scope-statistics}
\end{table}

\noindent Table~\ref{tab:scope-statistics} reports the utterance-distance distribution of gold emotion--cause pairs.
Most gold causes occur close to their target emotion utterances.
Specifically, 93.9\%, 87.0\%, and 93.7\% of the gold pairs in ECF, ConvECPE, and RECCON-DD, respectively, satisfy $0\leq i-j\leq4$.
Based on this consistent locality pattern, we use local scope as a empirical setting shared across datasets.
We also report results under full scope to evaluate the three paradigms.
Under local scope, gold pairs outside the candidate scope remain part of the evaluation and are counted as false negatives if not predicted.

\subsection{Competitive Baselines}

We compare the three controlled LLM paradigms with six representative ECPEC baselines.
\textbf{MRC}~\cite{liu2023} reformulates ECPEC as a three-stage machine reading comprehension process that sequentially identifies emotions, extracts their causes, and verifies candidate pairs.
A position-aware GCN models dialogue structure and speaker interactions.
\textbf{CENTER}~\cite{wang2024} detects center events and constructs event-aware graphs to model contextual regions and interactions among candidate emotion--cause pairs.
It jointly optimizes center event detection, emotion and cause extraction, and pair prediction.
\textbf{SCALE}~\cite{ma2026} models global contextual, local temporal, and intra-speaker dependencies using a shared conversation graph.
It derives emotion- and cause-oriented representations and aligns them through optimal transport.
\textbf{GMEC-GPT}~\cite{ju2025} enhances the generative GMEC framework by using GPT-3.5 to refine its predictions.
The refinement incorporates prediction confidence, retrieved examples, and visual descriptions generated by MiniGPT-4.
\textbf{DEC-Debate}~\cite{wu2026} combines an LLM-based generative model with a multimodal discriminative model.
Their conflicting emotion and pair predictions are resolved through multi-round LLM debates and majority voting.
\textbf{MSPF+}~\cite{xie2026a} performs pair verification by fusing tagging-based, synonym-based, and causal-claim prompt representations.
It further filters candidate pairs using a validation-selected utterance-distance threshold.

Since this work focuses on comparing LLM task paradigms rather than providing an exhaustive ECPEC leaderboard, we include only a compact set of representative baselines.
We therefore omit additional agent-based LLM systems, such as \textsc{HCPRA}~\cite{wang2026}, and task-specific architectures, such as \textsc{DAM}~\cite{kong2022}, \textsc{GSECE-UC}~\cite{an2023}, and \textsc{PRG-MoE}~\cite{jeong2023}.

\subsection{Reproducibility}

\paragraph{Training details.}
We use LLaMA-3.1-8B~\cite{llama3} as the primary backbone and further evaluate Qwen-2.5-7B~\cite{qwen2.5} and GLM-4-9B~\cite{glm4}.
For the paradigm-specific comparison, a separate LoRA adapter~\cite{hu2022lora} is fine-tuned for each backbone and task paradigm using the same training split.
All experiments use deterministic single-pass inference with temperature $0$.
For \textsc{Jud}, a candidate pair is predicted as positive when the normalized probability of the \texttt{Yes} token is greater than 0.5; ties are assigned to No. The threshold is fixed across datasets and is not tuned on validation or test data.
The training hyperparameters are summarized in Table~\ref{tab:training-hyperparameters}.

\begin{table}[h]
    \centering
    \scriptsize
    \setlength{\tabcolsep}{2pt}
    \begin{tabular*}{\columnwidth}{@{\extracolsep{\fill}}lccccccccc}
        \toprule
        & \multicolumn{3}{c}{\textbf{LLaMA-3.1-8B}}
        & \multicolumn{3}{c}{\textbf{Qwen-2.5-7B}}
        & \multicolumn{3}{c}{\textbf{GLM-4-9B}} \\
        \cmidrule(lr){2-4} \cmidrule(lr){5-7} \cmidrule(lr){8-10}
        \textbf{Hyperparameter}
        & \textsc{Gen} & \textsc{Sel} & \textsc{Jud}
        & \textsc{Gen} & \textsc{Sel} & \textsc{Jud}
        & \textsc{Gen} & \textsc{Sel} & \textsc{Jud} \\
        \midrule
        LoRA $r$ & 8  & 8 & 16 & 8 & 8 & 16 & 8 & 8 & 8 \\
        LoRA $\alpha$ & 16 & 16 & 32 & 16 & 16 & 32 & 16 & 16 & 16 \\
        Dropout & 0.00 & 0.00 & 0.05 & 0.00 & 0.00 & 0.05 & 0.00 & 0.00 & 0.05 \\
        Learning rate & 5e-5 & 5e-5 & 5e-5 & 5e-5 & 5e-5 & 5e-5 & 5e-5 & 5e-5 & 5e-5 \\
        Batch size & 1 (8) & 1 (8) & 1 (8) & 1 (8) & 1 (8) & 1 (8) & 1 (8) & 1 (8) & 1 (8) \\
        Epochs & 6 & 6 & 2 & 6 & 6 & 2 & 6 & 6 & 2 \\
        \bottomrule
    \end{tabular*}
    \caption{Hyperparameters for paradigm-specific fine-tuning.}
    \label{tab:training-hyperparameters}
\end{table}

\paragraph{Implementation.}
Our implementation uses Python 3.10.20, PyTorch 2.11.0, Transformers 5.9.0, PEFT 0.19.1, and CUDA 12.8.
All experiments are conducted on a server equipped with eight NVIDIA GeForce RTX 5090 GPUs (32 GB each) with two Intel Xeon Platinum 8457C CPUs and 1 TB of system memory.
We use a random seed of 13 and report results from a single run.

\paragraph{Evaluation metrics.}
Predictions from all three paradigms are converted into sets of ordered emotion--cause utterance-index pairs.
Duplicate pairs are removed, and indices outside the dialogue or the evaluated candidate scope are discarded.
A prediction is correct only when both utterance indices exactly match a gold pair.
Let $\mathcal{P}$ and $\mathcal{G}$ denote the predicted and gold pair sets over the complete test set.
We compute
\begin{equation}
    P=\frac{|\mathcal{P}\cap\mathcal{G}|}{|\mathcal{P}|},
    \quad
    R=\frac{|\mathcal{P}\cap\mathcal{G}|}{|\mathcal{G}|},
    \quad
    F_1=\frac{2PR}{P+R}.
\end{equation}
\section{Definition of Recoverable Omission Rates}\label{apx:recoverable-omission}

For each experimental condition, let $\mathcal{Y}$ denote the gold relation set.
We use the subscripts $G$, $S$, and $J$ to refer to \textsc{Gen}, \textsc{Sel}, and \textsc{Jud}, respectively, and let $\widehat{\mathcal{Y}}_G$, $\widehat{\mathcal{Y}}_S$, and $\widehat{\mathcal{Y}}_J$ denote their predicted relation sets.
The gold relations omitted by \textsc{Gen} are defined as
\begin{equation}
    FN_G = \mathcal{Y}\setminus\widehat{\mathcal{Y}}_G.
\end{equation}
The gold relations correctly identified by \textsc{Sel} and \textsc{Jud} are
\begin{equation}
    TP_S = \mathcal{Y}\cap\widehat{\mathcal{Y}}_S, \qquad
    TP_J = \mathcal{Y}\cap\widehat{\mathcal{Y}}_J.
\end{equation}

\paragraph{\textsc{Gen}$\rightarrow$\textsc{Sel}.}
The recoverable omission rate from \textsc{Gen} to \textsc{Sel} measures the proportion of gold relations omitted by \textsc{Gen} but correctly identified by \textsc{Sel}:
\begin{equation}
    \mathrm{RO}_{G\rightarrow S} = \frac{|FN_G\cap TP_S|}{|FN_G|}.
    \label{eq:gen-sel-ro}
\end{equation}

\paragraph{\textsc{Gen}$\rightarrow$\textsc{Jud}.}
The recoverable omission rate from \textsc{Gen} to \textsc{Jud} measures the proportion of gold relations omitted by \textsc{Gen} but correctly identified by \textsc{Jud}:
\begin{equation}
    \mathrm{RO}_{G\rightarrow J} = \frac{|FN_G\cap TP_J|}{|FN_G|}.
    \label{eq:gen-jud-ro}
\end{equation}

Both rates use the same set of \textsc{Gen} false negatives as the
denominator and are therefore directly comparable.
A higher rate indicates that a larger proportion of the relations omitted
by \textsc{Gen} remain identifiable under the corresponding finer-grained
paradigm.
All sets are aggregated over the test set before computing the rates.

\section{Use of AI Assistants}
AI assistants were used only for language polishing, formatting assistance, and consistency checking. 
All research ideas, experimental designs, implementations, analyses, and conclusions were developed and verified by the authors.

\end{document}